\documentclass[runningheads,a4paper]{llncs}
\usepackage{graphicx}
\usepackage{mathrsfs}
\usepackage{latexsym}
\usepackage{amssymb}
\usepackage{amscd}
\usepackage{amsmath}
\usepackage{url}
\usepackage{float}
\urldef{\mailsa}\path|{alfred.hofmann, ursula.barth, ingrid.haas, frank.holzwarth,|
\urldef{\mailsb}\path|anna.kramer, leonie.kunz, christine.reiss, nicole.sator,|
\urldef{\mailsc}\path|erika.siebert-cole, peter.strasser, lncs}@springer.com|
\newcommand{\keywords}[1]{\par\addvspace\baselineskip
\noindent\keywordname\enspace\ignorespaces#1}

\begin{document}

\mainmatter  

\title{Measuring Conceptual Entanglement in Collections of Documents}

\titlerunning{Conceptual Entanglement in Documents}

%

\author{Tom\'as Veloz$^{1,3}$, Xiaozhao Zhao$^2$, Diederik Aerts$^3$}
\authorrunning{} 
%
\tocauthor{}
\institute{
Department of Mathematics, \\
University of British Columbia, Kelowna, British Columbia, Canada\\
\email{tomas.veloz@ubc.ca} \\
\and
School of Computer Science \\
Tianjin University, Tianjin, China \\
\email{0.25eye@gmail.com} \\
\and
Center Leo Apostel (Clea) and Department of Mathematics,\\
Brussels Free University (VUB), Pleinlaan 2, 1050 Brussel, Belgium\\
\email{diraerts@vub.ac.be}
}

%
%

\toctitle{Entanglement in Documents}
\tocauthor{Veloz, Zhao, Hou, Aerts}
\maketitle
\vspace{-0.5cm}
\begin{abstract}
Conceptual entanglement is a crucial phenomenon in quantum cognition because it implies that classical probabilities cannot model non-compositional conceptual phenomena. While several psychological experiments have been developed to test conceptual entanglement, this has not been explored in the context of Natural Language Processing. In this paper, we apply the hypothesis that words of a document are traces of the concepts that a person has in mind when writing the document. Therefore, if these concepts are entangled, we should be able to observe traces of their entanglement in the documents.  In particular, we test conceptual entanglement by contrasting language simulations with results obtained from a text corpus. 
Our analysis indicates that conceptual entanglement is strongly linked to the way in which language is structured. We discuss the implications of this finding in the context of conceptual modeling and of Natural Language Processing. 
\vspace{-0.25cm}
\end{abstract}
\vspace{-1cm}
\section{Introduction\label{intro}}
\vspace{-0.25cm}
The extraction of relevant information from the web has been a fundamental area of research and development during the last decades~\cite{baeza1999modern}. In particular, several quantum-inspired approaches have been elaborated to perform standard tasks in Information Retrieval (see~\cite{zuccon2011use,wang2010tensor,piwowarski2009quantum,melucci2010investigation,melucci2011quantum,widdows2004geometry}, and references therein).\\ 
In order to extract information from the web it is necessary to establish a representation model. The most efficient techniques represent documents and terms as vectors in high dimensional spaces. These vectors are built based on statistical analyses of large corpora of web pages and documents. Examples of such methods include Latent Semantic Analysis~\cite{dumais2005latent}, Hyperspace Analogue to Language~\cite{lund1996producing}, and Latent Dirichlet Allocation~\cite{blei2003latent}.
While these approaches have been shown to be extremely useful for information retrieval (IR) tasks, they rely on that counting term occurrences is {\bf sufficient} to represent the meaning of a document in a corpus~\cite{wallach2006topic}. 
\\ 
Increasing evidence shows that a term-based analysis of a corpus of text is not sufficient to perform information extraction in an optimal way~\cite{baeza2011new,baeza2010chapter}. As an alternative to this term-based methodology, semantic-based models which emphasize the conceptual information that is represented by the terms in the document have been proposed~\cite{giunchiglia2009concept,fernandez2011semantically}. This means that the statistical information that we can obtain from the distribution of terms in a document can be complemented with conceptual information that we have in advance~\cite{kharkevich2010concept}.\\
The concept-based methods applied in Natural Language Processing (NLP) and IR are based mainly on the use of Ontologies~\cite{paralic2003ontology}, and Word-net like structures~\cite{fellbaum2010wordnet}. These approaches have been useful to improve syntactic approaches to NLP and IR. However, they cannot account for elements that have been identified as of primary interest in concept theory. Namely, the context-dependence of concepts' meaning~\cite{rosch1999principles}, and the non-compositional meaning of concept combinations~\cite{smith1984conceptual}. While Word-net and some ontology-based approaches consider context, it has been shown that it is computationally unfeasible to handle context-sensitivity in general~\cite{strang2004context,mandala1998use}. Moreover, none of both approaches considers the non-compositionality of concept combination.\\
To overcome these limitations, quantum-inspired models have been proposed~\cite{aerts2005theory,aerts2005theoryb}. In the quantum approach to concepts, the conceptual entity is assumed to be contextual. This means that its meaning emerges from its interaction with a certain context, analogous to how quantum states become actual after interacting with a measurement apparatus. Moreover, the mathematical formalism of Fock space allows to model the emergence of non-compositional meaning in conceptual combinations~\cite{aerts2009quantum,aerts2007general}. \\
Conceptual entanglement has been investigated to detect quantum behaviour in the phenomenology of concepts~\cite{bruza2008entangling,aerts2011quantum,aerts2013quantum,bruza2009extracting}. The discovery of conceptual entanglement has been crucial because it implies that a classical probability framework is not sufficient to describe the non-compositionality of conceptual phenomena~\cite{bruza2012quantum}. Particularly when a concept combination is entangled, a strong dependence between the terms that form instances of this concept combination is found. Such dependence is revealed by statistical tests, the so-called Bell-like inequalities~\cite{aerts2000violation}.\\
The aim of the present paper is to investigate whether conceptual entanglement is of significant importance on written texts. In particular, we view a piece of text as a trace of the concepts the subject who write had in mind at the while writing (for a complete elaboration of this perspective see~\cite{QI2013LSA}). From here, we assume that it is possible to measure conceptual entanglement by observing statistical properties of these conceptual traces.
A previous attempt to investigate this aspect is~\cite{aerts2010interpreting}. However, they analyze one concept combination only. We want to go one step further and investigate the extent to which entanglement can be found by automatic methods. If conceptual entanglement is significant on the web, then non-classical probabilities should become the {\it standard framework} for modeling in IR and NLP.\\
In section~\ref{Measuring-Entanglement} we explain how to measure entanglement using a co-occurrence corpus of text; in section~\ref{Method} we explain the methodology we used to analyze the corpus; and sections~\ref{Results} and~\ref{Discussion} illustrate the results and discuss their relevance.
\vspace{-0.35cm}
\section{Conceptual Entanglement}
\vspace{-0.25cm}
\label{Measuring-Entanglement}
\subsection{Detecting Conceptual Entanglement}
Following~\cite{aerts2000violation}, if we want to test whether two abstract entities $\cal A$ and $\cal B$ are entangled, we need to set two observables for each entity, each observable having two possible outcomes. Hence, we will denote these observables and their outcomes by $A=\{A_1,A_2\},A'=\{A'_1,A'_2\}$ for entity $\cal A$, and $B=\{B_1,B_2\},$ and $B'=\{B'_1,B'_2\}$ for entity $\cal B$. Next, we assume that 
each measurement $X\in\{A,A',B,B'\}$ corresponds 
the value $1$ if $X_1$ is observed and 
$-1$ if $X_2$ is observed. From here, we can define a composed experiment $XY\in\{AB,A'B,AB',A'B'\}$,   
corresponding to $1$ in case $X_1Y_1$ or $X_2Y_2$ is 
observed, and to $-1$ in case $X_1Y_2$ or $X_2Y_1$ is observed.\\
If we perform each experiment $XY$ a large number of times, we can estimate the expected value $E(XY)$ of each composed experiment. The Clauser-Horn-Shimony-Holt (CHSH) inequality states that if
\begin{equation}
-2\leq E(AB)+E(A'B)+E(AB')-E(A'B')\leq 2,
\label{CHSH}
\end{equation}
is not hold, then the entities are entangled. 
The violation of~\eqref{CHSH} implies the non-existence of one Kolmogorovian probability model for the considered joint experiments~\cite{accardi1982statistical}. \\
It is important to mention that the entities $\cal A$ and $\cal B$ need not be physical entities. The CHSH inequality is a statistical test that verifies whether or not it is possible to model a set of data in a classical probability setting.\\
For example in~\cite{aerts2011quantum}, the entities $\cal A$ and $\cal B$ refer to the concepts 
{\it Animal}  and 
{\it Acts}, respectively. The possible collapse states, i.e. the observables, of these entities were defined as 
$A=\{\text{`Horse',`Bear'}\}$, $A'=\{\text{`Tiger',`Cat'}\}$, and $B=\{\text{`Growls',`Whinnies'}\}$, and $B'=\{\text{`Snorts',`Meows'}\}$.
A psychological experiment where participants chose the combination that best represented the combination 
{\it The Animal Acts}, considering elements from $AB,~AB',~A'B$, and $A'B'$ was performed, and  the expected values of these joint observables were found to violate~\eqref{CHSH} with value 2.4197. Hence, it is concluded that the concepts $\cal A$ and $\cal B$ are entangled. 
\vspace{-0.25cm}
\subsection{Concepts and Meaning of a Document}
Someone who is writing a piece of text usually does not have the exact wording in mind, but only a particular idea of the intended meaning. If an idea is not concrete enough, it is hard to express it properly.
Usually in this case, the idea is broken into a set of interconnected concrete ideas. These concrete ideas are easier to put into sentences, which in turn form the paragraphs of the document. Moreover, it is only at the time of writing of a document that the words that express these concrete ideas are elicited. Hence, in this process we can identify two steps: an abstract idea is converted into several concrete ideas, and these concrete ideas are converted into sentences. Following this reflection, we propose a cognitive interpretation of the notion of meaning for pieces of text. More in particular, we assume that the process of converting an abstract idea into concrete ideas corresponds to the formation of ``entities of meaning'', and that these ``entities of meaning'' collapse to states which are represented by text, in the form of words or sentences. The meaning of a piece of text can thus be understood as the solution of an inverse problem, i.e. the piece of text plays the role of the collapsed state of an entity of meaning, a so-called conceptual trace, and the meaning of the piece of text is obtained by identifying the entity of meaning that collapsed to this piece of text.\\
In this work we do not focus on the inverse problem formulation as a general framework for meaning in documents. For an in-depth elaboration of this idea we refer to~\cite{QI2013LSA}. 
Instead, we assume the existence of concepts underlying the meaning of a document, and test whether or not these conceptual traces exhibit entanglement in text corpora.
\vspace{-0.25cm}
\subsection{Measuring Conceptual Entanglement in Text Corpora}
\label{measuring-entanglement-corpora}
Let $T$ be a corpus of text containing a set of terms $E$. Note that each term $t\in E$ can be considered an instance of one or many concepts. For example, the term 
{\it Dog} can be considered an instance of concepts 
{\it Pet}, 
{\it Animal}, 
{\it Mammal}, etc. Let  $C_1,C_2\subseteq E$ be sets of exemplars of concepts ${\cal C}_1$ and ${\cal C}_2$, respectively. 
Let $W$ be a positive integer. We say that concepts ${\cal C}_1$ and ${\cal C}_2$ $W$-co-occur if there exists a sequence $s$ of $W$ consecutive terms in the corpus $T$ such that one term in $C_1$ and one term in $C_2$ co-occur in $s$. We call $s$ a window of size $W$. In general, we can compute the $W$-co-occurrence frequency $
F_W({\cal C}_1,{\cal C}_2):C_1\times C_2 \to \mathbb{N}$ of exemplars of concepts ${\cal C}_1$ and ${\cal C}_2$ within windows of size $W$ in the corpus.\\
Suppose we choose $C_1$ and $C_2$ to have $4$ terms each, and we partition $C_1=\{A,A'\}$ and $C_2=\{B,B'\}$ such that each set in each partition has two terms. We then have that $A$ and $A'$ ($B$ and $B'$) are pairs of exemplars referring to concept $C_1$ ($C_2$). Hence, the co-occurrence measurements $AB,A'B,AB'$ and $A'B'$ can be used to measure the entanglement of concepts $C_1$ and $C_2$ in the corpus $T$. 
This is analogous to what has been done in previous psychological studies in conceptual entanglement~\cite{bruza2008entangling,bruza2009extracting,aerts2010interpreting,aerts2011quantum}.
However, in this work, instead of performing a psychological experiment where participants are requested to choose co-occurrent terms from a list, we extract these co-occurrences from a corpus of text.\\
For each pair $XY\in\{AB,A'B,AB',A'B'\}$, we compute their term co-occurrence
\begin{equation}
F_W(X_iY_j)=\sum_{s\in T} N(X_iY_j,s),
\label{freq-comp}
\end{equation}
where $i,j\in\{1,2\}$ and $N(X_i,Y_j,s)$ is equal to one if the pair $X_iY_j$ co-occurs in the window $s$ of the corpus $T$, and zero if not.\\
From here we can estimate the expected values of each co-occurrence experiment
\begin{equation}
E(XY)=\frac{F(X_1Y_1)+F(X_2Y_2)-F(X_1Y_2)-F(X_2Y_1)}{F(X_1Y_1)+F(X_2Y_2)+F(X_1Y_2)+F(X_2Y_1)},
\label{exp-val}
\end{equation}
Note that to measure conceptual entanglement in a corpus of text, we must first identify a set of exemplars for each concept. However, a multitude of concepts can be built upon a list of words, so that it is not possible to know in advance how to group words representing instances concepts. Indeed, psychological experiments where participants have to build categories choosing groups of words from a list show that, although some words fall regularly into similar kinds of categories, every word set can be potentially considered as a category~\cite{ranjan2010dissimilarity}.    \\
Instead of proposing a methodology to identify concepts, we assume that relevant words in a document correspond to relevant traces of the concepts that entail the document's meaning. Hence, without knowing exactly which concepts we are measuring entanglement for, we are able to represent them by sets of terms $C_1$ and $C_2$ from the relevant words of a document, following a given relevance criterion. 
Therefore, we propose the following {\it brute force} algorithm to measure entanglement: 
\begin{enumerate}
\item Select two sets $C_1$ and $C_2$ of relevant terms from the corpus, having $k$ words each. 
\item Verify the CHSH inequality for all the elements 
$(c_1,c_2)\in {\cal P}(C_1,4)\times {\cal  P}(C_2,4)$, where ${\cal P}(C,n)$ is the set of all the possible subsets of length $n$ of the set $C$.
\end{enumerate}
In order to perform the second step of the algorithm, we verify if there is a way of partitioning $c_1$ and $c_2$ such that~\eqref{CHSH} is violated. To do so, we consider all the possible row/column permutations of the $4\times 4$ matrix $F_W(c_1,c_2)$, and apply the expected co-occurrence formula~\eqref{exp-val} to estimate~\eqref{CHSH}.
Note that we have 12 different partitions for each $c_1$ and $c_2$, leading to 144 partitions for $(c_1,c_2)$. Hence, we say that $C_1$ and $C_2$ are entangled if there exists $(c_1,c_2)\in {\cal P}(C_1,4)\times {\cal  P}(C_2,4)$ such that (at least) one of the 144 co-occurrence matrices generated from $(c_1,c_2)$ violates the CHSH inequalities. 
Our aim is to estimate the likelihood of finding instances $(c_1,c_2)$ of concepts $C_1$ and $C_2$, such that they can be partitioned in a way that violates the CHSH inequalities. 
\vspace{-0.25cm}
\subsection{Statistical Considerations}
\label{statistical-considerations}
Consider the following question: Assume we know in advance the term co-occurrence frequency distribution of the corpus $T$, i.e. we know the probability $\rho_T(n)$ that two terms co-occur $n$ times in $T$ for each value of $n$;  what is the likelihood of building concepts $(c_1,c_2)\in {\cal P}(T,4)\times {\cal  P}(T,4)$ that violate the CHSH inequality in the corpus $T$?\\
In order to answer this question, we need to compare the kinds of co-occurrence matrix that $\rho_T(n)$ delivers, to the kinds of frequency matrix that violate the CHSH inequality.\\
Note that inequality~\eqref{CHSH} is likely to be violated if the three first elements have the same signs and relatively large values, and the fourth term has the opposite sign and a relatively large value. Indeed, in such case each row leads to an expected value near to $1$ (or -1), except for the fourth row, which leads to an expected value near to $-1$ (or 1). 
Table~\ref{F-matrix} shows an example of frequency data that violates the CHSH inequality with a positive value near $4$.
\vspace{-0.5cm}
\begin{table}[h!]
\vspace{-0.1cm}
\begin{center}
\begin{tabular}{c|c|c|c|c|c}
$F_W()$ & $B_1$ & $B_2$ & $B_1'$ &$B_2'$&  \\ \hline
$A_1$&L&S&L&S\\ \hline
$A_2$&S&L&S&L\\ \hline
$A_1'$&L&S&S&L\\ \hline
$A_2'$&S&L&L&S\\ \hline
\end{tabular}
\end{center}
\caption{Co-occurrence table that violates the CHSH inequality. The leftmost column and top row indicate terms of concepts $C_1$ and $C_2$, respectively. The interior $4\times 4$ table indicates the types of co-occurrence. $L$ stands for 'Large' and $S$ stands for 'Small'. If we switch columns $B_2/B_1$ and $B_2'/B_1'$, the inequality is violated with a negative value.}
\label{F-matrix}
\vspace{-0.75cm}
\end{table}

From table~\ref{F-matrix} we infer that if $\rho_T(n)$ does not have co-ocurrences that are relatively larger than others, then the likelihood of violating the CHSH inequality is low. 
Therefore, a corpus $T$ will be able to violate the CHSH inequality if $\rho_T(n)$ is significant for small values of $n$, and also for large values of $n$. This argument is qualitative and intuitive, but it can be better explained with an example. Suppose the corpus $T$ has a maximal co-occurrence equal to $100$, and the co-occurrences are divided into three groups of values: small values $G_S=\{1,...,5\}$, intermediate values $G_I=\{10,...,20\}$, and large values $G_L=\{50,...,100\}$. Note that, in case that all the co-occurrence frequencies belong to a single group, the expected value $E(XY)$ given by equation~\eqref{exp-val} is likely to be small. However, if one of the values belong to $G_L$ and all the other values belong to $G_S$, then $E(XY)$ will be dominated by the value in $G_L$. A list of other possible cases is shown in table~\ref{cooc-violation-cases}.
\vspace{-0.25cm}
\begin{table}[h!]
\vspace{-0.1cm}
\begin{center} 
\begin{tabular}{|c|c|c|c|c|c|c|c|c|}\hline
&Freqs&Expected val&&Freqs&Expected val& &Freqs&Expected val\\ \hline
1&SSSS&S& 5 &IIII&S&9    &LIII&L	 	\\ \hline
2&ISSS&S& 6 &LSSS&L&10   &LSSL&L		\\ \hline
3&ISSI&I&   7 &LSSI&L&  11   &LSIL&L		\\ \hline
4&ISII&S&   8 &LSII&L&     12  &LIIL&L		\\ \hline 
\end{tabular}
\end{center}
\caption{Co-occurrence frequencies and their expected values determined from equation~\eqref{exp-val}. Intermediate frequencies tend to decrease the likelihood of obtaining large expected values.}
\label{cooc-violation-cases}
\end{table}
\vspace{-1cm}
From table~\ref{cooc-violation-cases} we can see that intermediate co-occurrence values tend to diminish the likelihood if violating the CHSH inequality. Note that the question we analyze in this section assumes we can choose the partition of $c_1$ and $c_2$ that maximizes the evaluation of~\eqref{CHSH}. Indeed, table~\ref{cooc-violation-cases} considers the partition that maximizes the evaluation of~\eqref{CHSH} for each case.

\section{Method}
\label{Method}
\vspace{-0.2cm}
\subsection{Statistical Language Analysis}
\label{baseline}
We first computed the probability $p_B(\lambda)$ that, given a distribution  $\rho(n)$ of co-occurrences, there exists a partition for two random sets, having $4$ terms each, that violates the CHSH inequality. The distributon of co-occurrences $\rho(n)$ is modeled with a $B$-bounded Zipfian distribution $Z_B(\lambda,n)$ of parameter $\lambda$. We have chosen a Zipfian distribution because it has been shown to model the statistics of term co-occurrence in English language~\cite{i2001small}, and we bounded the distribution by a parameter $B$ to model the fact that the term co-occurrence frequency is limited by the corpus size. 
We also computed the likelihood of finding conceptual entanglement using homogeneous and Poisson distributions. These distributions do not exhibit significant entanglement so we do not present their results. However, we want to remark that this is consistent with the reasoning of section~\ref{statistical-considerations} because neither homogenous nor Poisson distributions assign large probabilities to both small and large values of $n$.\vspace{-0.25cm}

\subsection{The collection of documents}
We evaluate the aforementioned entanglement test on TREC collection: WSJ8792. Lemur 4.12 \footnote{Lemur is an open source project that develops search engines and text analysis tools for research and development of information retrieval and text mining softwares} is used for indexing. The collection is pre-processed by removing stop terms and applying the Porter stemmer. Among TREC topics 151-200, 32 topics with more than 70 truly relevant documents (judged by users) are selected. For certain topics, all truly relevant documents are segmented into windows of $W$ consecutive terms ($W=20,10,5,$ in our experiments). We considered two alternative ways to automatically generate concepts $C_1$ and $C_2$:
\begin{description}
  \item[$\Box$] Top 20 frequent terms $\{t_1^i,t_2^i,\dots,t_{20}^i\}$ of each topic $i=1,...,32$, are extracted from the set of truly relevant documents.
 Concept $C_1^i$ consists of top 10 frequent terms, i.e., $\{t_1^i,t_2^i,\dots,t_{10}^i\}$, and the rest of terms forms concept $C_2^i$.
  \item[$\Box$] Top 20 terms with largest tf-idf are extracted from the set of truly relevant documents, denoted as $\{{t'_1}^i,{t'_2}^i,\dots,{t'_{20}}^i\}$ (ranked by tf-idf values). Concept $C_1^i$ consists of top 10 terms, i.e., $\{{t'_1}^i,{t'_2}^i,\dots,{t'_{10}}^i\}$, and the rest of terms form concept $C_2^i$.
\end{description}
Then the windows are used to count the co-occurrence frequencies between exemplars from concepts $C_1^i$ and $C_2^i$ for each $i=1,...,32$.\vspace{-0.3cm}
\section{Results}
\label{Results}
\vspace{-0.25cm}
\subsection{Statistical Language Analysis}
\begin{figure}[h!]
\begin{center}
\includegraphics[height=3cm,width=12cm]{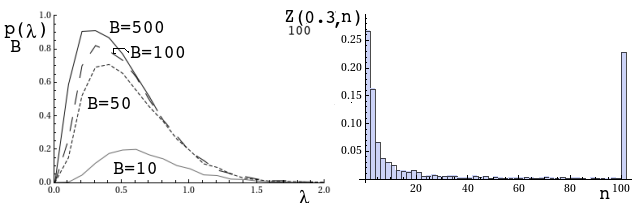}
\caption{Entanglement probability for random co-occurrence matrix on the left, and two examples of Zipfian distributions at the center and right, respectively.} 
\label{random-analysis}	
\end{center}
\end{figure}

Figure~\ref{random-analysis} shows the probability $p_B(\lambda)$ described in section~\ref{baseline}, for $0<\lambda\leq 2$, and $B=10,50,100,500,$ on the left. Note that for $\lambda\sim 0.3$, the proportion of entangled concepts is maximized for all values of $B$. Note that English co-occurrence distribution is estimated to be a Zipfian distribution of parameter $0.66\leq \lambda\leq0.8$~\cite{i2001small}. This range exhibits around $50\%$ of entanglement when $B\geq 50$. This indicates that when the size of the corpus is statistically significant, and hence co-occurrences are allowed to have large values, the likelihood of finding entangled concepts is non-neglectable. Therefore, we predict that entanglement is not a rare effect, and hence it should be observable from the {\it brute force} procedure proposed in section~\ref{measuring-entanglement-corpora}.\\
The middle and right plots in figure~\ref{random-analysis} show the Zipfian co-occurrence distributions $Z_B(\lambda,n)$ for $B=100$, using $\lambda=0.3$ and $\lambda=0.7$, respectively. These distributions confirm that when co-occurrence probability mass is concentrated on small and large values, conceptual entanglement is more likely to be detected. 
\vspace{-0.5cm}
\vspace{-1.5cm}
\subsection{Corpus Analysis}
\begin{figure}[h!]
\begin{center}
\includegraphics[height=4cm,width=12cm]{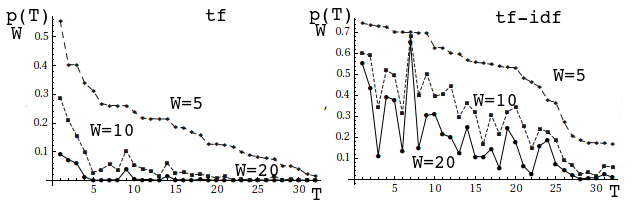}
\vspace{-0.2cm}
\caption{Entanglement probability for the 20 most frequent terms. The left plot corresponds to the co-occurrence data for most frequent terms, and the right plot corresponds to the co-occurrence data for the td-idf highest ranked terms.}
\label{corpus-analysis-pic}	
\end{center}
\vspace{-0.75cm}
\end{figure}

Figure~\ref{corpus-analysis-pic} shows the proportion $p_W(T)$ of entangled subsets of terms for the different topics in the corpus. The black curve corresponds to $W=20$, the gray curve corresponds to $W=10$ and the black dashed curve corresponds to $W=5$. The left plot is based on the data obtained using the frequency relevance method, and the right plot is based on the tf-idf method. In both cases, $p_W(T)$ is strictly decreasing with respect to $W$. Moreover, we observe that $p_W(T)$ does not have uniform variation with respect to $W$. For example in the tf-idf case (right plot in figure~\ref{corpus-analysis-pic}), $p_W(7)$ has a very close value for $W=5,10,$ and $20$. Analogously, $p_W(T)$ does not exhibit major variations for topics $1,10,16,19,$ and topics $24$ to $32$ for $W=20$ and $10$. However, $p_W(T)$ exhibits a major change when $W$ changes from $10$ to $5$ for most topics.    
Therefore, given two topics $T_1,T_2$, and two window sizes $W_1,W_2$, we cannot ensure that $p_{W_1}(T_1)<p_{W_1}(T_2)$ implies that $p_{W_2}(T_1)<p_{W_2}(T_2)$. Hence, there is not a topic-sorting such that $p_W(T)$ is a strictly decreasing function for $W=5,10,$ and $20$ simultaneously. Analogously, for the term-relevance method, while topics have a different set of relevant terms, there is not a unique sorting of the topics set that leads to $p_W(T)$ decreasing for $W=20,10$ and $5$ simultaneously. \\ 
In both methods, topics were sorted such that $p_5(T)$ is decreasing. By doing so, we avoid that curves cross each other because $p_W(T)$ is strictly decreasing with respect to $W$.
Note that $p_W(T)$ reaches significantly larger values for the tf-idf method. This is consistent with the fact that tf-idf is a better indicator of term relevance than the term frequency. 
\vspace{-0.5cm}
\subsection{Distribution of co-occurrences}
\begin{figure}[b!]
\vspace{-0.75cm}
\begin{center}
\includegraphics[height=5cm,width=11cm]{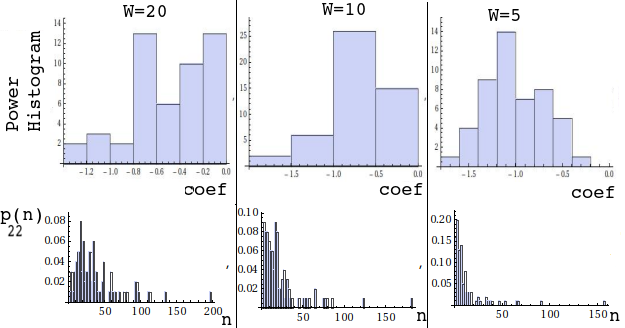}
\vspace{-0.5cm}
\caption{Co-occurrence distribution examples extracted from the tf-idf method. The first row corresponds to the third topic, and the second row corresponds to topic number 7. The left plot of each row corresponds to $W=22$, the center plot to $W=10$ and the right plot to $W=5$.}
\end{center}
\label{Distribution-analysis}	
\end{figure}

\vspace{-0.1cm}
We computed the histogram $\rho_T(n)$ for each topic $T$ in the corpus, and for each window size. In figure~\ref{Distribution-analysis}, we plot examples considering window sizes $W=20,10$ and $5$, for topics $3$ and $7$ of the co-occurrences data obtained from the tf-idf method.
\\
We observe that when the window size decreases, $\rho_T(n)$ tends to decrease the number of co-occurrences on intermediate values. However, large co-occurrence values remain probable. From here we infer that when the window size decreases, the co-occurrence distribution becomes similar to a bounded Zipfian distribution. Observe for example the first row of figure~\ref{Distribution-analysis}; we can see qualitative changes on the probability distribution for the different window sizes. This is consistent with the strong changes observed in $p_W(T=3)$ of figure~\ref{corpus-analysis-pic} for the tf-idf method. Analogously, the second row does not exhibit major changes in its co-occurrence distribution. This is again reflected in $p_W(T=3)$ of figure~\ref{corpus-analysis-pic}, where no major changes are also observed. 
This is consistent with the fact that the $p_W(T)$ is strictly decreasing with respect to $W$. 
\vspace{-0.5cm}
\section{Conclusion}
\vspace{-0.2cm}
\label{Discussion}
We performed a qualitative analysis to corroborate that conceptual entanglement is a significant effect in written texts. The analysis is based on the hypothesis that pieces of text are traces of concepts, and that such concepts entail the meaning of documents~\cite{QI2013LSA}.
This corroboration is supported by an analysis of statistical properties of the English language, and of 32 topic-structured text corpora.  
Our language analysis indicates that the statistical co-occurrence distribution of the English language has a significant tendency to build conceptual entities that are entangled. Particularly, term co-occurrence matrices built from a distribution that models co-occurrence of words in English, violate the CHSH inequality in around $50\%$ of the cases. It should be noted that this analysis is based only on the statistical distribution of co-occurrences of English, and does not consider any semantic or linguistic aspect of the English language.\\
We further analyzed a corpus separated by 32 topics, and found results consistent with the language analysis. For each topic, we built a matrix computing the term co-occurrence of the 10 most relevant terms with respect to the next 10 most relevant terms, considering window sizes $W=5,10$, and $20$ to measure co-occurrence. From here, we computed the proportion of $4\times 4$ sub-matrices of the co-occurrence matrix that violate the CHSH inequality. We observe that the tf-idf relevance delivers more entanglement than the term frequency relevance measure. This is consistent with the fact that td-idf is a better relevance measure than term frequency. Although some topics exhibit more entanglement than others, we identify a strong tendency to find conceptual entanglement for most topics. Moreover, conceptual entanglement decreases with respect to the window size. This is consistent with the fact that word correlations are noisy for large window sizes~\cite{i2001small}. For short window sizes, we have more chance to keep only meaningful correlations, and hence entanglement is observed with more clarity.
In addition, we found that for shorter window sizes, the distribution of co-occurrences becomes more similar to a bounded Zipfian distribution. Indeed, in figure~\ref{corpus-analysis-pic}, we can see that when the distributions look more like a Zipfian distribution, i.e. when $W=5$, the average conceptual entanglement for words selected by the tf-idf relevance measure is $50\%$ of entanglement. \\
A fundamental and novel element of this work is that we do not build in advance the concepts for which we will measure entanglements. Instead, we assume that relevant words of a topic-corpus are relevant traces of the concepts that entail the meaning of the topic~\cite{QI2013LSA}. Hence, if the co-occurrence of these traces violates the CHSH inequality, we conclude that the concepts that entail the meaning of the document are entangled. Therefore, we do not focus on the generation of categories or taxonomies. However, we also suggest that one interesting extension of this work would consist in testing conceptual entanglement from categories that are automatically built as in~\cite{roark1998noun,widdows2002graph}.
We also propose to study why some topics exhibit more entanglement than others, count term co-occurrence in structured sentences, e.g. Frame-Net~\cite{baker1998berkeley}, rather than in windows of text, and evaluate other statistical conditions that provide a more precise classification of entanglement~\cite{marginal-select}.  

\end{document}